\documentclass[sigconf]{acmart} 

\usepackage{amssymb}
\usepackage{amsmath,mathtools}
\usepackage{multirow}
\AtBeginDocument{%
  }
\providecommand{\doi}[1]{\href{https://doi.org/#1}{doi: \nolinkurl{#1}}}

\setcopyright{none}
\acmConference[Preprint]{Preprint}{September 2026}{}
\acmBooktitle{Preprint}
\acmDOI{}
\acmISBN{}

\newcommand{\auxpctr}{auxiliary pCTR tower}
\newcommand{\extpctr}{serving pCTR model}
\newcommand{\logextpctr}{logged serving pCTR}
\newcommand{\logextpctrscore}{\logextpctr{} score}
\newcommand{\logextpctrscores}{\logextpctr{} scores}
\newcommand{\esmmdcn}{ESMM+DCN-V2}
\makeatletter
\newcommand{\namedversiononly}[1]{\if@ACM@anonymous\else#1\fi}
\makeatother

\begin{document}
\raggedbottom


\title[Serving-Aligned Entire-Space Conversion Modeling]
{Serving-Aligned Entire-Space Conversion Modeling via Click-Through Rate Knowledge Distillation}

\author{Pavel Dolin}
\authornote{Work performed while at Reddit.}
\affiliation{%
  \institution{Arizona State University}
  \city{Tempe}
  \state{AZ}
  \country{United States}}

\author{Xun Zou}
\affiliation{%
  \institution{Reddit}
  \city{San Francisco}
  \state{CA}
  \country{United States}}

\author{Jiatian Wu}
\authornote{Corresponding author: \href{mailto:alex.wu@reddit.com}{\nolinkurl{alex.wu@reddit.com}}.}
\affiliation{%
  \institution{Reddit}
  \city{San Francisco}
  \state{CA}
  \country{United States}}

\author{Deguang Kong}

\renewcommand{\shortauthors}{Dolin et al.}

\begin{abstract}
Recommendation systems may use predicted click-through conversion rate (pCTCVR), the product of predicted click-through rate (pCTR) and post-click conversion rate (pCVR), to guide ranking decisions. Many pCVR models build on the Entire Space Multi-Task Model (ESMM), which jointly trains pCVR and auxiliary pCTR towers using pCTR and pCTCVR losses, but deploys only the pCVR tower. At serving time, this tower may be paired with a separately trained pCTR model, creating a training-serving mismatch in pCTCVR when its predictions differ from those of the auxiliary pCTR tower.
In this paper, we use knowledge distillation (KD) from logged serving pCTR scores to supervise ESMM's auxiliary pCTR tower and improve training-serving pCTCVR alignment.
We denote variants by KD--$\alpha$, where $\alpha$ controls the hard-label loss weight. On the Ali-CCP dataset, KD--0.3, the best variant on serving pCTCVR, increases ROC-AUC not only for serving pCTCVR but also for pCVR by 2.1\% and 1.6\%, respectively, and PR-AUC by 9.9\% and 8.8\%, respectively, relative to the ESMM baseline.
We also test negative downsampling with 5\% nonclick retention. Downsampled KD--0.3 increases mean AUC for serving pCTCVR and pCVR by 1.2\% and 0.9\%, respectively, and PR-AUC by 8.5\% and 4.5\%, respectively, relative to downsampled ESMM.
\end{abstract}



\maketitle

\begin{figure}[h]
    \centering
    \captionsetup{skip=10pt}
    \includegraphics[width=0.9\columnwidth]{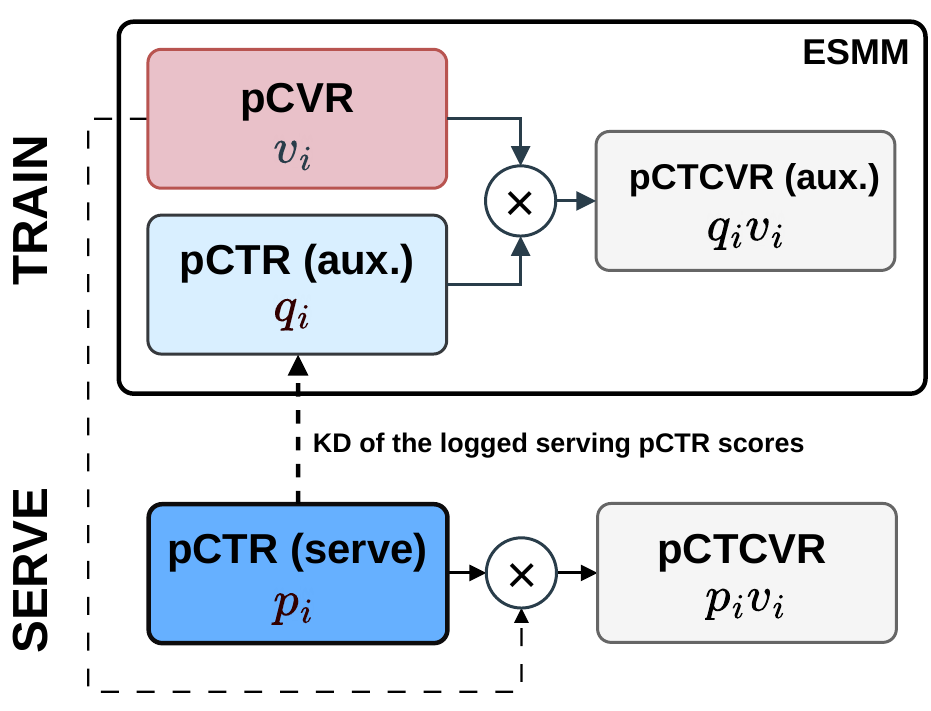}
    \caption{Full-population KD. ESMM forms auxiliary pCTCVR $q_i v_i$ and uses
    the \logextpctrscore{} $p_i$ to supervise $q_i$; serving combines $p_i$
    with $v_i$ to form $p_i v_i$.}
    \Description{Two vertically arranged training and serving paths. In the upper
    training path, an ESMM box contains an auxiliary pCTR tower producing $q_i$
    and a pCVR tower producing $v_i$; their scores enter a multiplication node
    that produces auxiliary pCTCVR $q_i v_i$. In the lower serving path, the
    logged serving pCTR score $p_i$ is multiplied by the trained pCVR score $v_i$
    to produce serving pCTCVR $p_i v_i$. A dashed arrow from $p_i$ to the
    auxiliary pCTR tower indicates that $p_i$ supervises $q_i$ through knowledge
    distillation.}
    \label{fig:method}
\end{figure}

\section{Introduction}
Recommendation systems rank items based on predicted user actions, such as clicks and purchases, modeled through predicted click-through rate (pCTR) and post-click conversion rate (pCVR), respectively. For conversion-oriented ranking, the utility score may depend on predicted click-through conversion rate (pCTCVR), computed as the product of pCTR and pCVR scores from separately trained models \citep{obrien2021analysis}.

Click feedback is fast, whereas conversion events, such as purchases, may occur days or weeks later \citep{chapelle2014delayed}. Conversion feedback is also substantially sparser than click feedback \citep{obrien2021analysis,wen2020esm2}. These differences can motivate separate training windows and update schedules \citep{yang2021capturing}. Moreover, click prediction is used across multiple stages of ranking systems \citep{gao2025supply}, providing additional motivation for independent pCTR model development. This separation is particularly relevant to ESMM and its variants. ESMM jointly trains pCVR and auxiliary pCTR towers using pCTR and pCTCVR losses \citep{ma2018esmm}, but deploys only the pCVR tower, which may be paired with an independently trained \extpctr{}. Differences between the auxiliary and serving pCTR predictions can therefore create a training-serving mismatch in the resulting pCTCVR score.

We address this mismatch from the pCVR side rather than by modifying the \extpctr{}. Production pCTR models can be large and computationally expensive to train \citep{anil2022factory}. Jointly training them with pCVR models may be costly and operationally impractical, particularly when pCTR predictions are shared across ranking components. Instead, we use knowledge distillation (KD) \citep{hinton2015distilling} to supervise ESMM's \auxpctr{} with \logextpctrscores{} from the frozen \extpctr{}. This improves training-serving alignment without retraining the serving pCTR model or modifying ESMM's architecture (Figure~\ref{fig:method}). We further demonstrate the effectiveness of our method under negative downsampling, a common practice in production ranking systems to address class imbalance and reduce training costs \citep{he2014practical,liu2024at4ctr}. The paper makes the following contributions:
\begin{itemize}
  \item \textbf{Method.} We distill \logextpctrscores{} into ESMM's auxiliary pCTR tower to align training and serving pCTCVR predictions (Figure~\ref{fig:method})\namedversiononly{\footnote{Code is available at \url{https://github.com/izuminka/esmm_kd_pctr}}}. 

  \item \textbf{Full-population results.} On Ali-CCP \cite{ma2018esmm}, KD--0.3 improves mean serving pCTCVR and pCVR AUC by 2.1\% and 1.6\%, PR-AUC by 9.9\% and 8.8\%, and reduces NE by 0.4\% and 0.6\%, respectively, over ESMM (Section~\ref{sec:results-full-population}).

  \item \textbf{Negative-downsampling results.} With 5\% nonclick retention, KD$_{\mathcal R}$--0.3 improves mean serving pCTCVR and pCVR AUC by 1.2\% and 0.9\%, PR-AUC by 8.5\% and 4.5\%, and reduces NE by 0.3\% for both, over ESMM$_{\mathcal R}$ (Section~\ref{sec:results-downsampled}).


\end{itemize}

\section{Related Work}

\paragraph{Independent pCTR--pCVR composition}
Ma et al. \citep{ma2018esmm} evaluate pCTCVR by combining pCVR predictions with a common independently trained pCTR model. O'Brien et al. \citep{obrien2021analysis} study independent and multi-task approaches to conversion modeling, while Beeraka et al. \citep{beeraka2025ipcf} examine the evaluation of composed ranking models. 


\paragraph{Entire-space conversion modeling}
ESMM jointly learns pCTR and pCVR by supervising pCTR and pCTCVR over all
impressions \citep{ma2018esmm}. ESM$^2$, AITM, and ESCM$^2$ extend this approach
with post-click state decomposition, sequential transfer, and counterfactual
objectives \citep{wen2020esm2,xi2021aitm,wang2022escm2}.
UKD and EVI distill CVR predictions to extend conversion
supervision to unclicked impressions, mitigating sample
selection bias and data sparsity
\citep{xu2022ukd,fei2025evi}.
In contrast, our method distills serving pCTR scores
into ESMM's auxiliary pCTR tower to align the pCTR
factors used in training and serving.

\paragraph{Knowledge Distillation}
Wang et al. \citep{wang2023ranking} distill final-stage CTR predictions into an early-stage ranking model to improve ranking consistency. Yang et al. \citep{yang2022crosstask} use cross-task distillation to transfer knowledge between recommendation objectives, while Xu et al. \citep{xu2020privileged} distill models using privileged features unavailable at serving time. In contrast, we distill \logextpctrscores{} into ESMM's auxiliary pCTR tower to align the pCTR factors used in training and serving pCTCVR computation.



\section{Method}
\label{sec:method}



\subsection{Setup}

Let $X$ denote impression features and let $Y,Z$ denote binary click and click-and-conversion labels, with $Z\leq Y$. For observations $(x_i,y_i,z_i)_{i=1}^N$, define the full set of impressions $\mathcal U=\{1,\ldots,N\}$ and the downsampled subset of impressions $\mathcal R=\{i\in\mathcal U:I_i=1\}$, where $I$ is the retention indicator and $I_i$ is its value for impression $i$. ESMM outputs auxiliary pCTR $q_i=q(x_i)$ and pCVR $v_i=v(x_i)$. A frozen \extpctr{} $P$ provides $p_i=\operatorname{stopgrad}(P(x_i))$ as supervision, not an input feature. Auxiliary pCTCVR is $q_i v_i$; serving uses $p_i v_i$. For binary or soft targets $a$ and predicted probabilities $b$, binary
cross-entropy is $\ell(a,b)=-a\log b-(1-a)\log(1-b)$. Write $\langle g\rangle_{\mathcal A}=|\mathcal A|^{-1}
\sum_{i\in\mathcal A}g_i$ for the average over a nonempty index set $\mathcal A$.

\subsection{Full-Population Training}
\label{sec:method-full-space}

The ESMM baseline jointly optimizes the auxiliary pCTR and pCTCVR losses:
\begin{equation}
\mathcal L_{\mathrm{ESMM}}^{\mathcal U}
=\left\langle
  \ell(y_i,q_i)+\ell(z_i,q_i v_i)
 \right\rangle_{\mathcal U}.
\label{eq:method-esmm}
\end{equation}
KD replaces the auxiliary pCTR loss with a weighted combination of hard-label and teacher supervision, while leaving the pCTCVR loss unchanged. For teacher scores $t_i\in[0,1]$ and weight $\alpha\in[0,1]$, define
\begin{equation}
\ell_{\mathrm{KD}}(y_i,t_i;q_i)
=\alpha\ell(y_i,q_i)+(1-\alpha)\ell(t_i,q_i).
\label{eq:method-kd-row}
\end{equation}
Here, $\alpha=1$ recovers the ESMM baseline, while $\alpha=0$ uses only teacher supervision for auxiliary pCTR.
Using the full-population teacher score $t_i=p_i$ gives
\begin{equation}
\mathcal L_{\mathrm{KD}}^{\mathcal U}
=\left\langle
  \ell_{\mathrm{KD}}(y_i,p_i;q_i)+\ell(z_i,q_i v_i)
 \right\rangle_{\mathcal U}.
\label{eq:method-full-kd-full}
\end{equation}

The Direct baseline removes the \auxpctr{} and supervises the
serving product using the \logextpctrscore{} $p_i$:
\begin{equation}
\mathcal L_{\mathrm{Direct}}^{\mathcal U}
=\left\langle\ell(z_i,p_i v_i)\right\rangle_{\mathcal U}.
\label{eq:method-direct}
\end{equation}

\subsection{Training with Negative Downsampling}
\label{sec:method-downsampled-space}

We retain all clicks and sample each nonclick independently with probability
$\beta\in(0,1]$: $\Pr(I=1\mid X,Y,Z)=1$ for $Y=1$ and $\beta$ otherwise.
The teacher scores $p_i$ are unchanged by downsampling and still refer to
the full population. Distillation on the retained population therefore
requires a correction. The retained-space baseline, denoted ESMM$_{\mathcal R}$, uses
an unweighted loss on $\mathcal R$:
\begin{equation}
\mathcal L_{\mathrm{ESMM}}^{\mathcal R}
=\left\langle
  \ell(y_i,q_i)+\ell(z_i,q_i v_i)
 \right\rangle_{\mathcal R}.
\label{eq:method-retained-esmm}
\end{equation}

\paragraph{Teacher correction}
For a full-population calibrated teacher, $\Pr(Y=1\mid P(X)=p)=p$.
Retaining all clicks and a fraction $\beta$ of nonclicks gives, by Bayes' rule,
\begin{equation}
\tau_\beta(p)
\coloneqq\Pr(Y=1\mid P(X)=p,I=1)
=\frac{p}{p+\beta(1-p)}.
\label{eq:method-tau-rho}
\end{equation}
With imperfect teacher calibration, this correction is approximate
\citep{dalpozzolo2015calibrating}. We denote retained-space KD by
KD$_{\mathcal R}$--$\alpha$: the subscript marks training on $\mathcal R$ using the
corrected teacher target $\tau_\beta(p_i)$. Its objective is
\begin{equation}
\mathcal L_{\mathrm{KD}}^{\mathcal R}
=\left\langle
  \ell_{\mathrm{KD}}\!\left(y_i,\tau_\beta(p_i);q_i\right)
  +\ell(z_i,q_i v_i)
 \right\rangle_{\mathcal R}.
\label{eq:method-retained-kd-r}
\end{equation}

\paragraph{Evaluation}
On complete validation and test splits, we map the auxiliary pCTR back to the
full-population scale using $q_i^{\mathrm{full}}\coloneqq\tau_\beta^{-1}(q_i)
=\beta q_i/(1-q_i+\beta q_i)$, correcting for the higher click prevalence in the
downsampled training set.
We use $q_i^{\mathrm{full}}$ for auxiliary pCTR and $q_i^{\mathrm{full}}v_i$ for
auxiliary pCTCVR; serving remains $p_i v_i$. Only $q_i$ is transformed. Exact
pCTR recovery requires $q_i$ to match the retained click posterior.

\section{Experimental Setup}
\label{sec:experimental-setup}

\paragraph{Data and Training.}
We keep the official Ali-CCP\footnote{\url{https://tianchi.aliyun.com/dataset/408}} \cite{ma2018esmm}
train/test split and hold out about 10\% of training impressions for validation, giving 38.07M training, 4.23M validation, and 43.02M test impressions. We train a Deep \& Cross Network V2 (DCN-V2) model \citep{wang2021dcnv2} to generate the \logextpctrscores{}. For pCVR baselines, we use ESMM and \esmmdcn{}. ESMM uses shared feature embeddings and separate multilayer perceptron (MLP) towers for auxiliary pCTR and pCVR prediction. \esmmdcn{} replaces the auxiliary pCTR MLP tower with DCN-V2 while retaining joint training. We use proportion-preserving stratified minibatches and initialize output biases to empirical positive-rate log-odds. We train each student model with 10 random seeds, selecting checkpoints by validation pCTCVR NE (serving for KD/Direct, auxiliary for ESMM variants), with the first epoch selected for nearly all runs.

\paragraph{Evaluation}
We report  area under the receiver operating characteristic (ROC) curve (AUC), area under the precision--recall curve (PR-AUC), and normalized entropy (NE), the mean log loss divided by binary label entropy. pCTR and pCTCVR use all test impressions; pCVR uses clicked impressions.
Auxiliary pCTR predictions are inverse-corrected as described in
Section~\ref{sec:method-downsampled-space} when training is downsampled.
For score alignment probabilities $a,b$, we use Bernoulli divergence in nats
$d(a\|b)=a\log\frac{a}{b}+(1-a)\log\frac{1-a}{1-b}$,
with boundary values defined by limits. We report the mean Kullback--Leibler
(KL) divergences $\mathrm{KL}(a\|b)=\langle d(a\|b)\rangle_{\mathrm{test}}$,
alongside Pearson correlations $r(a,b)$.
Evaluation metrics for each student model are averaged over ten seeds, except
metrics for the \logextpctrscores{}.
Relative-change 95\% confidence intervals (CIs) use Fieller's method on paired seeds.

\begin{table}[h]
\centering
\small
\setlength{\tabcolsep}{2pt}
\renewcommand{\arraystretch}{0.95}
\begin{tabular*}{\columnwidth}{@{\extracolsep{\fill}}clrrr@{}}
\hline
Model & Output & AUC $\uparrow$ & PR-AUC $\uparrow$ & NE $\downarrow$ \\
\hline
DCN-V2 & $p_i$ & 0.6310 & 0.06461 & 0.9815 \\
\hline
ESMM & $q_i$ & 0.6259 & 0.06328 & 0.9822 \\
\hline
\esmmdcn{} & $q_i$ & 0.6299 & 0.06486 & 0.9804 \\
\hline
KD--0.3 & $q_i$ & 0.6301 & 0.06432 & 0.9786 \\
\hline
\end{tabular*}
\caption{Full-population Ali-CCP pCTR test means. 
$q_i$ averaged over ten student seeds, and $p_i$ is the single seed.
}
\label{tab:aliccp-full-ctr}
\end{table}

\begin{figure}[t]
\centering
\captionsetup{skip=2pt}
\includegraphics[width=\columnwidth]{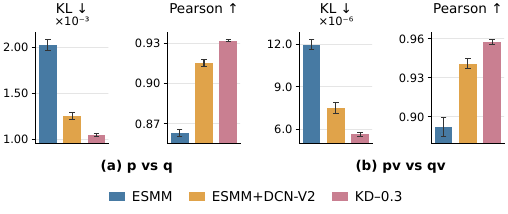}
\caption{Full-population score alignment for (a) logged serving versus
auxiliary pCTR ($p$ vs. $q$) and (b) serving versus auxiliary pCTCVR ($pv$ vs.
$qv$). Lower KL and higher Pearson $r$ indicate closer alignment;
bars show means $\pm$ standard error of the mean (SEM) over ten seeds.}
\Description{Four bar charts compare ESMM, \esmmdcn{}, and KD--0.3. In panel
(a), $\mathrm{KL}(p\|q)$ decreases and $r(p,q)$ increases from ESMM to
\esmmdcn{} to KD--0.3. In panel (b), $\mathrm{KL}(pv\|qv)$ likewise decreases
and $r(pv,qv)$ increases in the same order. KD--0.3 therefore has the lowest
divergence and highest correlation in both comparisons. Error bars show one
SEM, and the vertical axes are truncated.}
\label{fig:ctr-alignment}
\end{figure}

\section{Results and Discussion}
\label{sec:results}


\subsection{Full-Population Training}
\label{sec:results-full-population}

%

\begin{table}[h]

\centering
\small
\setlength{\tabcolsep}{2pt}
\renewcommand{\arraystretch}{1.0}
\begin{tabular*}{\columnwidth}{@{\extracolsep{\fill}}clrrr@{}}
\hline
Model & Output & AUC $\uparrow$ & PR-AUC $\uparrow$ & NE $\downarrow$ \\
\hline
\multirow{2}{*}{ESMM} & $v_i$ & 0.6949 & 0.01310 & 0.9616 \\
 & $p_i v_i$ & 0.6599 & 0.0004588 & 0.9867 \\
\hline
\multirow{2}{*}{\esmmdcn{}} & $v_i$ & 0.7027 & 0.01387 & 0.9591 \\
 & $p_i v_i$ & 0.6679 & 0.0004746 & 0.9854 \\
\hline
\hline
\multirow{2}{*}{KD--0.7} & $v_i$ & 0.6956 & 0.01311 & 0.9625 \\
 & $p_i v_i$ & 0.6635 & 0.0004668 & 0.9873 \\
\hline
\multirow{2}{*}{KD--0.5} & $v_i$ & 0.7007 & 0.01360 & 0.9590 \\
 & $p_i v_i$ & 0.6675 & 0.0004818 & 0.9841 \\
\hline
\multirow{2}{*}{KD--0.3} & $v_i$ & 0.7063 & 0.01425 & 0.9558 \\
 & $p_i v_i$ & 0.6736 & 0.0005042 & 0.9825 \\
\hline
\multirow{2}{*}{KD--0} & $v_i$ & 0.7033 & 0.01447 & 0.9565 \\
 & $p_i v_i$ & 0.6732 & 0.0005007 & 0.9848 \\
\hline
\multirow{2}{*}{Direct} & $v_i$ & 0.6973 & 0.01367 & 0.9610 \\
 & $p_i v_i$ & 0.6658 & 0.0004633 & 0.9861 \\
\hline
\end{tabular*}
\caption{Full-population Ali-CCP test results averaged over ten student seeds. Here, $v_i$ denotes pCVR, $p_i$ the single-seed \logextpctrscore{}, and $p_i v_i$ serving pCTCVR. pCVR is evaluated on clicked impressions, while serving pCTCVR is evaluated on all impressions.}
\label{tab:aliccp-full-seeds-column}
\end{table}


\paragraph{KD improves serving pCTCVR}
As the hard-label weight $\alpha$  decreases from 0.7 to 0.3, serving pCTCVR AUC and PR-AUC improve, while NE decreases. Among the tested KD--$\alpha$ settings, KD--0.3 performs best on serving
$p_i v_i$, with the highest AUC and PR-AUC and the lowest NE. With 95\%
paired-seed CIs in brackets, the relative serving AUC and PR-AUC gains over
ESMM are 2.1\% [1.7\%, 2.5\%] and 9.9\% [7.7\%, 12.3\%], and NE decreases
by 0.4\% [0.3\%, 0.5\%]. Against \esmmdcn{}, AUC and PR-AUC increase by
0.9\% [0.5\%, 1.2\%] and 6.2\% [$-0.3$\%, 12.8\%], and NE decreases by
0.3\% [0.1\%, 0.5\%]. Against Direct, the respective changes are 1.2\%
[0.7\%, 1.6\%], 8.8\% [1.7\%, 16.2\%], and a 0.4\% [0.3\%, 0.5\%] NE reduction
(Table~\ref{tab:aliccp-full-seeds-column}). Notably, KD--0, teacher supervision only, performs only marginally worse on pCTCVR metrics than KD--0.3.

\paragraph{KD aligns both pCTR and pCTCVR scores}
Figure~\ref{fig:ctr-alignment} provides evidence for the alignment mechanism
behind KD--0.3's gains. Against \esmmdcn{}, $\mathrm{KL}(p\|q)$ falls by
16.3\% and $\mathrm{KL}(pv\|qv)$ by 24.9\%; both correlations increase.
Aligning $q_i$ with $p_i$ reduces the mismatch between the training pCTCVR $q_i v_i$ and the serving pCTCVR $p_i v_i$. Table~\ref{tab:aliccp-full-ctr} reinforces this interpretation:
the \logextpctrscore{} $p_i$ and the auxiliary pCTR score $q_i$ have similar aggregate
pCTR performance, and KD--0.3 and \esmmdcn{} have comparable AUC and PR-AUC. The gain therefore reflects closer alignment with the
specific \logextpctrscores{} rather than a uniformly stronger auxiliary pCTR ranker.

\paragraph{KD outperforms baselines on pCVR AUC, PR-AUC, and NE}
The gains are not limited to serving pCTCVR. Relative to ESMM, KD--0.3
increases mean pCVR AUC and PR-AUC by 1.6\% [1.4\%, 1.9\%] and
8.8\% [7.6\%, 10.0\%], while reducing NE by 0.6\% [0.5\%, 0.7\%]. Against
\esmmdcn{}, AUC and PR-AUC improve by 0.5\% [0.2\%, 0.8\%] and
2.8\% [1.1\%, 4.5\%], with a 0.3\% [0.2\%, 0.5\%] NE reduction. Compared
to Direct, the respective improvements are 1.3\% [1.0\%, 1.6\%],
4.2\% [2.7\%, 5.8\%], and a 0.5\% [0.3\%, 0.7\%] NE reduction
(Table~\ref{tab:aliccp-full-seeds-column}). KD--0 has an estimated 1.56\%
[$-0.03$\%, 3.16\%] higher mean pCVR PR-AUC than KD--0.3. These results
suggest that aligning the auxiliary pCTR with the serving pCTR can improve
pCVR prediction, even without directly modifying the pCVR loss.



\subsection{Training with Negative Downsampling}
\label{sec:results-downsampled}


\paragraph{Ranking gains persist at 5\% nonclick retention}
Against ESMM$_{\mathcal R}$, KD$_{\mathcal R}$--0.3 raises
mean AUC for serving pCTCVR and pCVR by 1.2\% [0.9\%, 1.4\%] and
0.9\% [0.7\%, 1.1\%], respectively. PR-AUC increases by
8.5\% [2.7\%, 14.4\%] and 4.5\% [3.6\%, 5.4\%], respectively, while NE
decreases by 0.3\% [0.2\%, 0.5\%] and 0.3\% [0.2\%, 0.4\%], respectively
(Table~\ref{tab:aliccp-downsampled-5pct-column}). The gains on
both tasks thus persist with sampled negatives, and all four mean NE values are below one.


\begingroup
\setlength{\intextsep}{6pt}
\begin{table}[!htbp]

\centering
\small
\setlength{\tabcolsep}{2pt}
\renewcommand{\arraystretch}{1.0}
\begin{tabular*}{\columnwidth}{@{\extracolsep{\fill}}clrrr@{}}
\hline
Model & Output & AUC $\uparrow$ & PR-AUC $\uparrow$ & NE $\downarrow$ \\
\hline
\multirow{2}{*}{ESMM$_{\mathcal R}$} & $v_i$ & 0.6888 & 0.01248 & 0.9635 \\
 & $p_i v_i$ & 0.6540 & 0.0004514 & 0.9886 \\
\hline
\multirow{2}{*}{KD$_{\mathcal R}$--0.3} & $v_i$ & 0.6949 & 0.01305 & 0.9607 \\
 & $p_i v_i$ & 0.6617 & 0.0004900 & 0.9855 \\
\hline
\end{tabular*}
\caption{Ali-CCP test means over ten student seeds after training with 5\%
nonclick retention. Here, $v_i$ is the pCVR score, $p_i$ is the single seed
\logextpctrscore{}, and $p_i v_i$ is the
serving pCTCVR score. pCVR is evaluated on clicked impressions; serving pCTCVR
uses all test impressions.}
\label{tab:aliccp-downsampled-5pct-column}
\end{table}
\endgroup



\section{Limitations}
\label{sec:limitations}
Our evaluation focuses on the core alignment method using ESMM. Future work could extend the evaluation to more complex ESMM variants and additional datasets.

\section{Conclusion}
We presented a knowledge distillation (KD) approach that aligns ESMM's auxiliary pCTR tower with the \extpctr{} without modifying its architecture. On Ali-CCP, KD--0.3 improves both serving pCTCVR and pCVR predictive performance, increasing AUC by 2.1\% and 1.6\%, and PR-AUC by 9.9\% and 8.8\%, respectively, relative to ESMM. These gains persist under negative-downsampled training with 5\% nonclick retention.

\bibliographystyle{ACM-Reference-Format}
\bibliography{references}

\end{document}